\documentclass[final]{cvpr}

\usepackage{times}
\usepackage{epsfig}
\usepackage{graphicx}
\usepackage{amsmath}
\usepackage{amssymb}

\usepackage{bbm}
\usepackage{booktabs}
\usepackage{multirow}
\usepackage{pifont}

\usepackage[pagebackref=true,breaklinks=true,colorlinks,bookmarks=false]{hyperref}

\def\cvprPaperID{1391} 
\def\confYear{CVPR 2021}

\begin{document}

\title{Jigsaw Clustering for Unsupervised Visual Representation Learning}

\author{Pengguang Chen$^{1}$ \quad\quad Shu Liu$^{2}$ \quad\quad Jiaya Jia$^{1,2}$  \\[0.2cm]
	The Chinese University of Hong Kong$^{1}$\quad SmartMore$^{2}$\\
	\{pgchen, leojia\}@cse.cuhk.edu.hk \quad liushuhust@gmail.com
}

\maketitle
\pagestyle{empty}
\thispagestyle{empty}

\begin{abstract}

Unsupervised representation learning with contrastive learning achieved great success. This line of methods duplicate each training batch to construct contrastive pairs, making each training batch and its augmented version forwarded simultaneously and leading to additional computation. We propose a new jigsaw clustering pretext task in this paper, which only needs to forward each training batch itself, and reduces the training cost. Our method makes use of information from both intra- and inter-images, and outperforms previous single-batch based ones by a large margin. It is even comparable to the contrastive learning methods when only half of training batches are used.
 
Our method indicates that multiple batches during training are not necessary, and opens the door for future research of single-batch unsupervised methods.
Our models trained on ImageNet datasets achieve state-of-the-art results with linear classification, outperforming previous single-batch methods by 2.6\%. Models transferred to COCO datasets outperforms MoCo v2 by 0.4\% with only half of the training batches. Our pretrained models outperform supervised ImageNet pretrained models on CIFAR-10 and CIFAR-100 datasets by 0.9\% and 4.1\% respectively. Code is available at \url{https://github.com/Jia-Research-Lab/JigsawClustering}

\end{abstract}

\section{Introduction}
Unsupervised visual representation learning, or self-supervised learning, is a long-standing problem, which aims at obtaining general feature extractors without human supervision. This goal is usually achieved by carefully designing pretext tasks without annotation to train feature extractors.

\begin{figure}
	\centering
	\includegraphics[width=\linewidth]{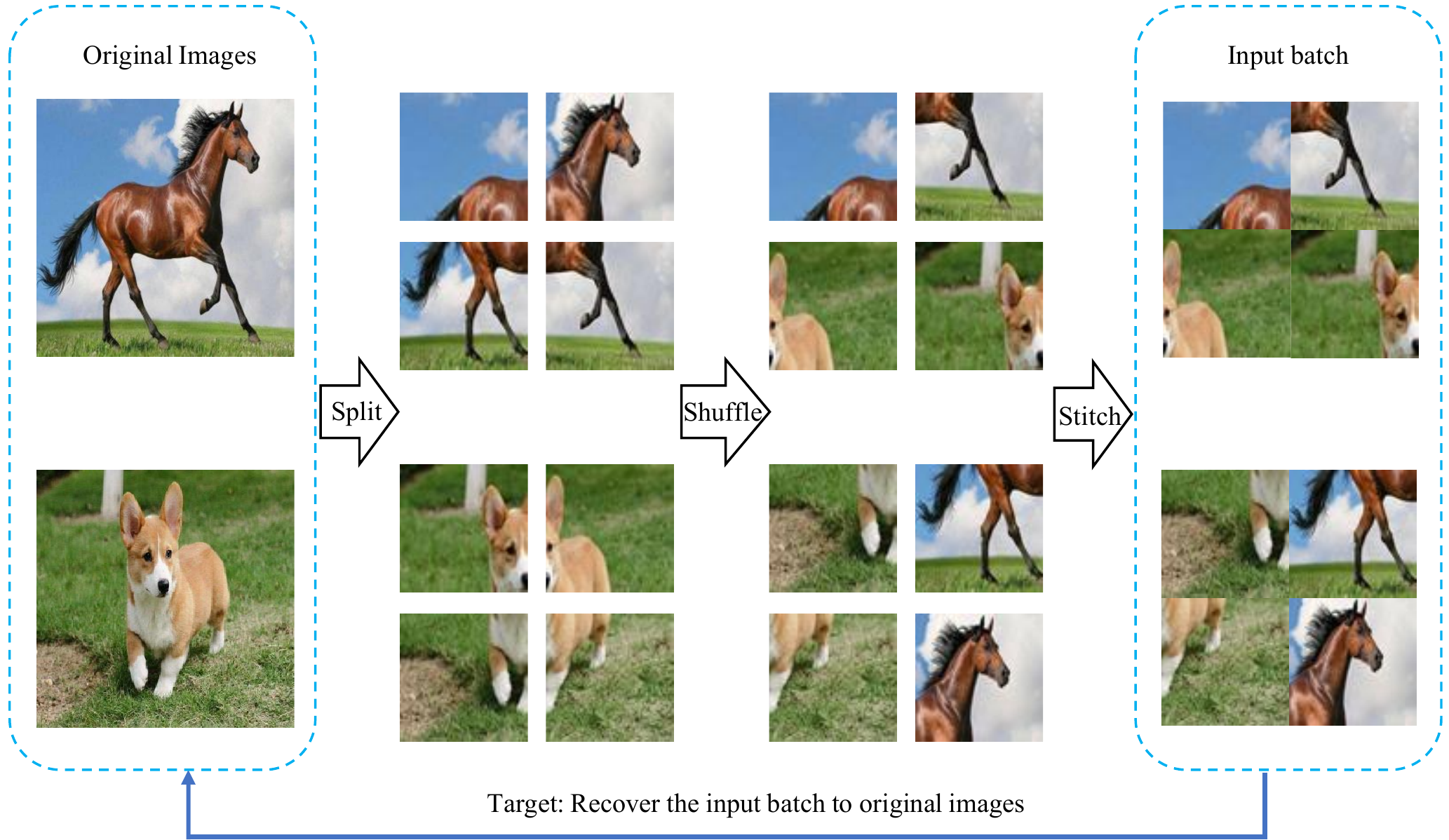}
	\caption{Sketch of our proposed pretext task. Images in the same batch are split into multiple patches, which are shuffled and stitched to form a new batch as input images for the network. The target is to recover the batch similar to the original images. We use two images here as an example.}
	\label{fig:teaser}
	\vspace{-0.1in}
\end{figure}

According to the definition of pretext tasks, most mainstream approaches fall into two classes: intra-image tasks and inter-images tasks. Intra-image approaches, including colorization \cite{colorization,autocolor} and jigsaw puzzle \cite{jigpuz}, design a transform of one image and train the network to learn the transform. Since only the training batch itself is forwarded each time, we name them {\it single-batch methods}. This kind of tasks can be achieved using only one image's information, limiting the learning ability of feature extractors. 

Inter-images tasks are developed rapidly in recent years, which require the network to discriminate among different images. Contrastive learning is popular now since it reduces the distance between representation of positive pairs and enlarges the distance between representation of negative pairs. To construct positive pairs, another batch of images with different augmented views are used in the training process \cite{simclr,moco,pirl}. Since each training batch and its augmented version are forwarded simultaneously, we name these methods {\it dual-batches methods}. They greatly raise resource required for training an unsupervised feature extractor. The way to design an efficient single-batch based method with similar performance to dual-batches methods is still an open problem.

In this paper, we propose a framework for efficient training of unsupervised models using Jigsaw Clustering (JigClu). Our method combines advantages of solving jigsaw puzzles and contrastive learning, and makes use of both intra- and inter-image information to guide feature extractor. It learns more comprehensive representations. Our method only needs a single batch during training and yet greatly improves results compared to other single-batch methods. It even achieves comparable results with dual-batches methods with only {\it half} of the training batches. 

\paragraph{Jigsaw Clustering Task} In our proposed Jigsaw Clustering task, every image in a batch is split into different patches. They are randomly permuted and stitched to form a new batch for training. The goal is to recover these disrupted parts back to the original images, as shown in Figure \ref{fig:teaser}. Different from \cite{jigpuz}, the patches are permuted in a batch instead of a single image. The image each patch belongs to and the location of each patch in the origin are predicted in our work. 

Also, we use montage images instead of single patches as input of the network.
This modification greatly improves the difficulty for the task of \cite{jigpuz} and provides more useful information for the network to learn. The network now has to distinguish between different parts of one image and identifies their original positions to recover the original image from multiple montage input images. 

This task allows the network to learn both intra- and inter-images information by only forwarding the stitched images, using half of the training batches compared to other contrastive learning methods. 

\begin{figure*}
	\centering
	\includegraphics[width=\linewidth]{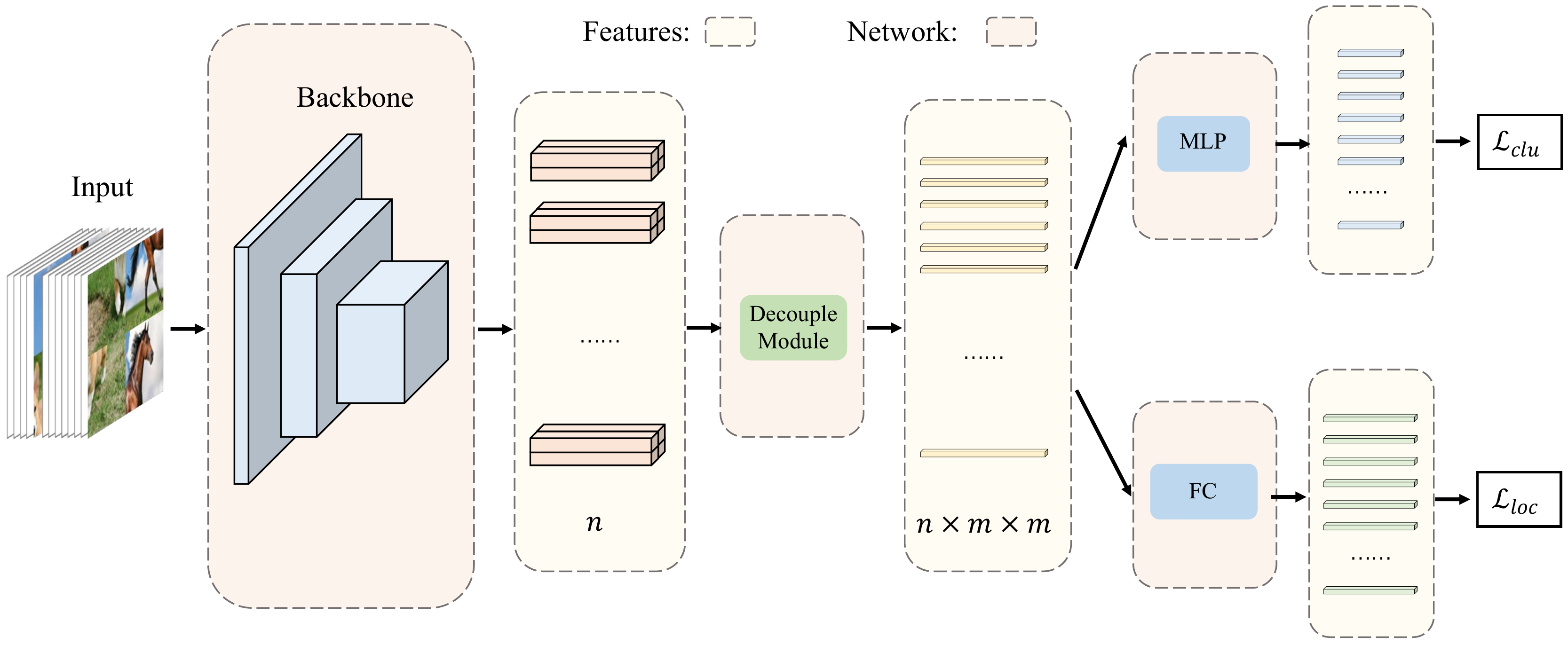}
	\caption{Pipeline of our method. We use light yellow rectangles to represent features produced by different parts of the network and light pink rectangles to represent parts of the network. The input images first go through the backbone network to produce $n$ feature maps. Then the $n$ feature maps are decoupled into $n\times m\times m$ vectors, each corresponding to one patch through a parameter-free decouple module. Afterwards, a MLP and a FC are used to embed vectors into logits to compute clustering loss and localization loss separately. }
	\label{fig:pipeline}
\end{figure*}

To recover patches across images, we design a clustering branch and a location branch as shown in Figure \ref{fig:pipeline}. Specifically, we first decouple the global feature map of stitched images into the representation of each patch. Then these two branches operate on representation of each patch. The clustering branch is to separate these patches into clusters, each of which only contains patches from the same image. The location branch, on the other hand, predicts location of every patch in an image agnostic manner. 

With prediction from these two branches, the Jigsaw Clustering problem is solved. The clustering branch is trained as a supervised clustering task since we know the patches are from the same image, or not. The location branch is considered as a classification problem, where each patch is assigned with a label to indicate its location in the origin image. This branch predicts the label of every patch. 

The reason that our method achieves decent results is that models trained with our proposed task can learn different kinds of information. At first, discriminating among different patches in one stitched image forces the model to capture instance-level information inside an image. This level of feature is missing in general in other contrastive learning methods.

Further, clustering different patches from multiple input images helps the model learn image-level features across images. This is the key that recent methods \cite{moco,mocov2,simclr} achieve high-quality results. Our method retain this important property.
Finally, arranging every patch to the correct location requires detailed location information, which was considered in single-batch methods \cite{jigpuz,colorization} before. It is, however, ignored in recent methods of \cite{simclr,pirl,moco,mocov2,pcl}. We note this piece of information is still important to further improve results. 

\paragraph{Performance of Our Method} Learning by our method yields both intra- and inter-images information. This comprehensive learning brings a spectrum of superiority. First, with only one batch during training, our method outperforms other single-batch ones by 2.6\% on linear evaluation on the ImageNe-1k dataset.
Second, our method is more data-efficient. When the training data size is not large, our method can still produce decent results, much better than many other existing ones. On the ImageNet-100 and ImageNet-10\% datasets, our system outperforms MoCo v2 by 6.2\% and 6.0\% respectively.
Our method also converges more quickly with less training time. We use only a quarter of epochs of MoCo v2 to achieve the same results on the ImageNet-100 dataset.

Finally, the comprehensive information learned by our models is suitable for many other vision tasks. On the COCO detection dataset, our result is 0.4\% better than MoCo v2, with only half of training batches. On the CIFAR-10 and CIFAR-100 datasets, models tuned with our pretrained weights achieve 0.9\% and 4.1\% higher results than that with supervised training weights, respectively.
The extensive experiments demonstrate the superiority of our proposed pretext method.

\section{Related Work}

\paragraph{Handcrafted pretext tasks} Many pretext tasks were proposed to train unsupervised models. Recovering the input image under corruption is an important topic, with tasks of descriminating synthetic artifacts \cite{spotartifacts}, colorization \cite{autocolor,colorization}, image inpainting \cite{inpainting}, and denoising auto-encoders \cite{denoiseautoencoder}, etc.
Besides, many methods generate persuade-labels by transformation to train the network without human annotations. Applications involve predicting relation of two patches \cite{contextpredict,transtive}, solving jigsaw puzzles \cite{jigpuz,damagedjigsaw}, and discriminating among surrogate classes \cite{exampler}. \cite{icbssl} is an improved vision of jigsaw puzzles \cite{jigpuz}, which utilizes more complex methods to choose patches.
Video information is also widely used for training unsupervised models \cite{moving,flowsimilarity,shufflelearn,watchmove}. 

\paragraph{Contrastive learning} Our method is also related to contrastive learning, which is first proposed in \cite{invariantmapping}. Following work \cite{dufec,npid,la,cmc} further improved performance. Recently, constructing contrastive pairs using different augmentation of images \cite{moco,simclr,pirl,sslhg} achieves great success. Espectially, \cite{sslhg} also utilize both intre- and inter-image information from pixel level. 
We note much training resource is required for training contrastive learning methods with multiple batches of images. Our work tackles this problem with newly designed contrastive pairs in a single batch.

\section{Jigsaw Clustering}
In this section, definition of the task is presented. We then propose a very simple network, which only needs hardly modification of the original backbone network, to accomplish this task. Finally, a novel loss function is designed to better suit our clustering task.
 
\subsection{The Jigsaw Clustering Task}
There are $n$ randomly selected images in one batch $\mathbf{X} = {x_1, x_2, ... , x_n}$. Every image $x_i$ is split into $m \times m$ patches. There are $n\times m \times m$ patches in a batch totally. These patches are randomly permuted to form a new batch of montage images $\mathbf{X}' = {x'_1, x'_2, ..., x'_n}$. Every new image consists of $m\times m$ patches, which may come from different images in $\mathbf{X}$.
 
The task is to cluster the $n\times m \times m$ patches given the new batch $\mathbf{X}'$ into $n$ clusters, and predict the location to recover the $n$ original images with every $m\times m$ patches in the same cluster. The process is shown in Figure \ref{fig:teaser}.

The key to the proposed task is to use montage images as input instead of every single patch. It is noteworthy that directly using small patches as input leads to the solution with only global information. Besides, small-size input images are not common for many applications. Only use them here raises the image-resolution difference problem between pretext and other downstream tasks. This may also lead to degradation of performance. Trivially scaling up small patches would violently increase the resource for training. 

Our montage images as input nicely avoid these drawbacks. First, the input images form only one batch with the same size as the original batch, which costs half of resource during training compared with recent methods \cite{simclr,moco}. More importantly, to better complete this task, the network has to learn detailed intra-image features to discriminate among different patches in one image, as well as global inter-image features to pull together different patches from the same original image. We observe that learning of comprehensive features greatly accelerates training of feature extractors. More experimental results are presented in Section \ref{sec:exp}.

\begin{figure}
	\vspace{-0.2in}
	\centering
	\includegraphics[width=\linewidth]{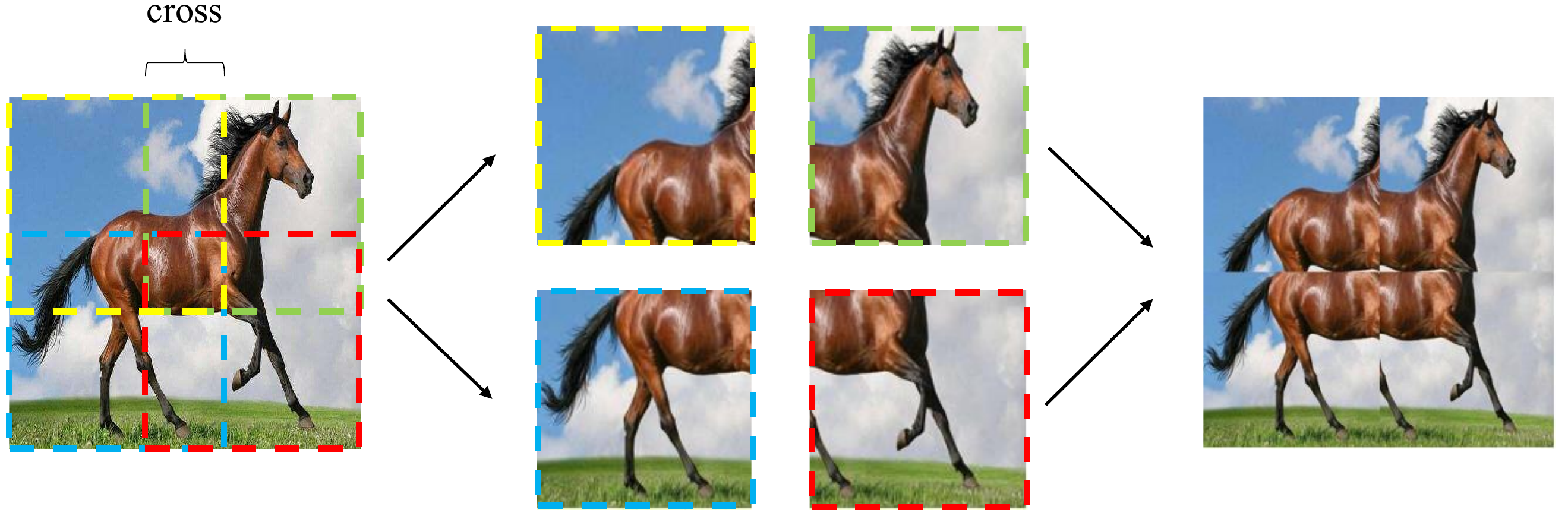}
	\caption{Patches split in images have a level of overlap.}
	\label{fig:cross}
	\vspace{-0.1in}
\end{figure}

The way to divide the image is a crucial part of our method. The choice of $m$ affects the difficulty of the task. Our ablation study on a subset of ImageNet (see Section \ref{sec:abl}) shows that $m=2$ achieves the best result. We conjecture that a larger $m$ would exponentially increase the complexity and make the network fail to learn effectively. Besides, we observe that cutting the image into disjoint patches is not optimal. With an extend of intersection as shown in Figure \ref{fig:cross}, the network learns better features. It is explainable that different regions of some images are too diverse. They cause difficulty for learning without any evidence of overlap. More analysis is presented in Section \ref{sec:abl}.

\subsection{Network Design}
We design a new decouple network for this task as illustrated in Figure \ref{fig:pipeline}. One module is a feature extractor, which can be any common architectures \cite{resnet,vgg,wideresnet,resnext,inceptionv2}. There is also a parameter-free decouple module to separate the feature into $m\times m$ parts corresponding to different patches in one input image. Then a multi-layer perceptron (MLP) is used to embed every patch's feature for the clustering task; a fully-connected layer (FC) is used for the localization task.

The decouple module first interpolates the feature map of the backbone into a new one whose side length is a multiple of $m$. We enlarge the feature map instead of narrowing it to avoid information loss. For example, a typical input size of ImageNet dataset is $224 \times 224$. The feature map produced by a ResNet-50 backbone is $7\times 7$. For $m=2$, we interpolate the feature map into $8\times 8$ by bilinear interpolation. When the length of the feature map is a multiple of $m$, we use average pooling to downsample the feature map to $n\times m \times m \times \hat{c}$. Then the features of a batch are disentangled to $(n\times m \times m) \times \hat{c}$, which means there are $(n\times m \times m)$ vectors of dimension $\hat{c}$. 

Every vector is then embedded into length $c$ with a two-layer MLP to form a set of vectors $\mathbf{Z} = {\boldsymbol{z}_1, \boldsymbol{z}_2, ..., \boldsymbol{z}_{nmm}}$ for the clustering task. In the meantime, a FC layer is attached after the $(n\times m \times m) \times \hat{c}$ vertors as the classifier to produce logits $\mathbf{L} = {\boldsymbol{l}_1, \boldsymbol{l}_2, ..., \boldsymbol{l}_{nmm}}$ for the localization task.

Our network is notably efficient, the additional decouple module is parameter-free. Compared to recent work, the computation of taking one batch remains almost the same, and we only need one batch during training. This greatly reduces training cost.

\subsection{Loss Functions}
\label{sec:loss}
The clustering branch is a supervised clustering task, because $m\times m$-size patches are in the same class. The supervised clustering task is convenient, and we use constractive learning to achieve it. We consider the target of clustering as pulling together objects from the same class and pushing away patches from different classes. Cosine similarity is used to measure the distance between patches. So for evey pair of patches in the same cluster, the loss function is
\begin{align}
\ell_{i,j} = - log \frac{\exp(cos(\boldsymbol{z}_i,\boldsymbol{z}_j)/\tau)}{\Sigma_{k=1}^{nmm}\mathbbm{1}_{k\neq i} \exp(cos(\boldsymbol{z}_i,\boldsymbol{z}_k)/\tau)},
\end{align}
where $\mathbbm{1}$ denotes the indicator function and $\tau$ is a temparture parameter to smooth or shappen the distance. The final loss function is summarized over all pairs from the same cluster as
\begin{align}
\mathcal{L}_{clu} = \frac{1}{nmm}\Sigma_{i}\left( \frac{1}{mm-1}\Sigma_{j\in \mathbf{C}_i}\ell_{i,j} \right),
\end{align}
where $\mathbf{C}_i$ denotes the set of patch indices in the same cluster of $i$.

The location branch is considered as a classification task. The loss function is simply cross-entropy loss, and the loss of localization is formulated as
\begin{align}
\mathcal{L}_{loc} = CrossEntropy(\mathbf{L}, \mathbf{L}_{gt}),
\end{align}
where $\mathbf{L}_{gt}$ denotes the ground truth for every-patch location.

The final objective of our proposed Jigsaw Clustering task is to optimize
\begin{align}
\mathcal{L} = \alpha \mathcal{L}_{clu} + \beta \mathcal{L}_{loc},
\end{align}
where $\alpha$ and $\beta$ are hyperparameters to balance these two tasks. In our experiments, $\alpha=\beta=1$ produces reasonable results.

\section{Experiments}
\label{sec:exp}

We report the performance of our unsupervised training method on ImageNet-1k \cite{imagenet} and ImageNet-100 datasets.

\paragraph{ImageNet-1k} is a widely used classification dataset. There are 1.2+ million images uniformly distributed in 1,000 classes. We use the training set without labels to train our models. 

\paragraph{ImageNet-100} is a subset of ImageNet-1k dataset, which is introduced in \cite{cmc}. This dataset randomly chooses 100 classes of ImageNet-1k, containing around 0.13 million images. It is also well balanced in terms of class distribution. We use this small dataset to verify data efficiency of our method and perform fast ablation studies.

\paragraph{Unsupervised Training} We use SGD to optimize our network with momentum 0.9. The weight decay is set to be $1e-4$. We train all models using batch size 256 on four GPUs. The learning rate is initialized as 0.03 and is decayed with cosine policy. All models are trained for 200 epochs if there is no further explanation.

\begin{table}
	\centering
	\begin{tabular}{c c c}
		\toprule
		Method & \# of Batch in Training & Accuracy \\
		\midrule
		Supervised & single-batch & 77.2 \\ 
		\midrule
		Colorization \cite{colorization} & single-batch & 39.6 \\
		JigPuz \cite{jigpuz} & single-batch & 45.7 \\
		DeepCulster \cite{deepcluster} & single-batch & 48.4 \\
		NPID \cite{npid} & single-batch & 54.0 \\
		BigBiGan \cite{bigbigan} & single-batch & 56.6 \\
		LA \cite{la} & single-batch & 58.8 \\
		SeLa \cite{sela} & single-batch & 61.5 \\ 
		CPC v2 \cite{cpcv2} & single-batch & 63.8 \\
		JigClu (Ours) & single-batch & \textbf{66.4} \\
		\midrule
		MoCo \cite{moco} & dual-batches & 60.6 \\
		PIRL \cite{pirl} & dual-batches & 63.6 \\
		SimCLR \cite{simclr} & dual-batches & 64.3 \\
		PCL \cite{pcl} & dual-batches & 65.9 \\
		MoCo v2 \cite{mocov2} & dual-batches & 67.7 \\
		\bottomrule
	\end{tabular}
	\vspace{0.1in}
	\caption{Linear evaluation results of ResNet-50 models on the ImageNet-1k dataset. Our model outperforms previous single-batch methods by a large margin, achieving comparable results with dual-batches methods.}
	\label{tab:linear-1k}
\end{table}

\subsection{Linear Evaluation}
We first evaluate the feature learned by our method with a linear classification protocol. 
We train a ResNet-50 backbone on the ImageNet dataset with unsupervised learning. Then a supervised linear classifier is trained on the top of the fixed backbone. The linear evaluation results on ImageNet-1k dataset are summarized in Table \ref{tab:linear-1k}. Our method outperforms previous single-batch based methods by a large margin, greatly reduces the gap from dual-batch based methods.

\paragraph{Comparison with JigPuz} 
JigPuz \cite{jigpuz} also solves the jigsaw puzzle for unsupervised learning. It defines the problem as sorting the patches inside every single image. Our method, contrarily, solves the jigsaw problem from a general perspective, and enrichs the feature learned from it. The Jigsaw Clustering task outperforms JigPuz by 19.9\% on the linear evaluation pipeline.

\paragraph{Comparison with Clustering Methods} 
DeepCluster \cite{deepcluster} and SeLa \cite{sela} are also based on clustering. But they use unsupervised clustering to guide the learning of models. We, instead, split images into different patches to generate ground truth for the clustering tasks. The supervised clustering task is more powerful for learning for our task and leads to much better representation. 

\paragraph{Comparison with Contrastive-based Methods} 
SimCLR \cite{simclr} and MoCo \cite{moco,mocov2} are recently proposed based on contrastive learning. They achieve high-quality results at the cost of more training resource, since an additional batch is required during training. These methods need to scan twice of the batches compared with single-batch based methods. We also utilize contrastive loss for our clustering task, but do not need extra batches. 

Our method achieves comparable results with state-of-the-art dual-batch based methods with only half of the training batches. MoCo v2 models yield slightly better results than ours on linear evaluation. Since MoCo v2 learns more of the inter-image information, it is suitable for the classification tasks. In contrast, our models learn comprehensive information, and therefore outperforms MoCo v2 on the detection tasks as shown in Section \ref{sec:coco}.

\paragraph{Data Efficiency}
We also experiment with our method on ImageNet-100 and a subset of ImageNet, which contains 10\% data of every class in ImageNet.  We train the ResNet-50 model on these datasets with unsupervised methods first. We report the linear evaluation results on the dataset to represent model ability. The results are presented in Table \ref{tab:linear2}, notably better than those of other contrastive learning methods on relatively small datasets. This is because our method makes use of both intra- and inter-image information. The comprehensive learning strategy utilizes limited data more effectively.

\begin{figure}
	\centering
	\includegraphics[width=\linewidth]{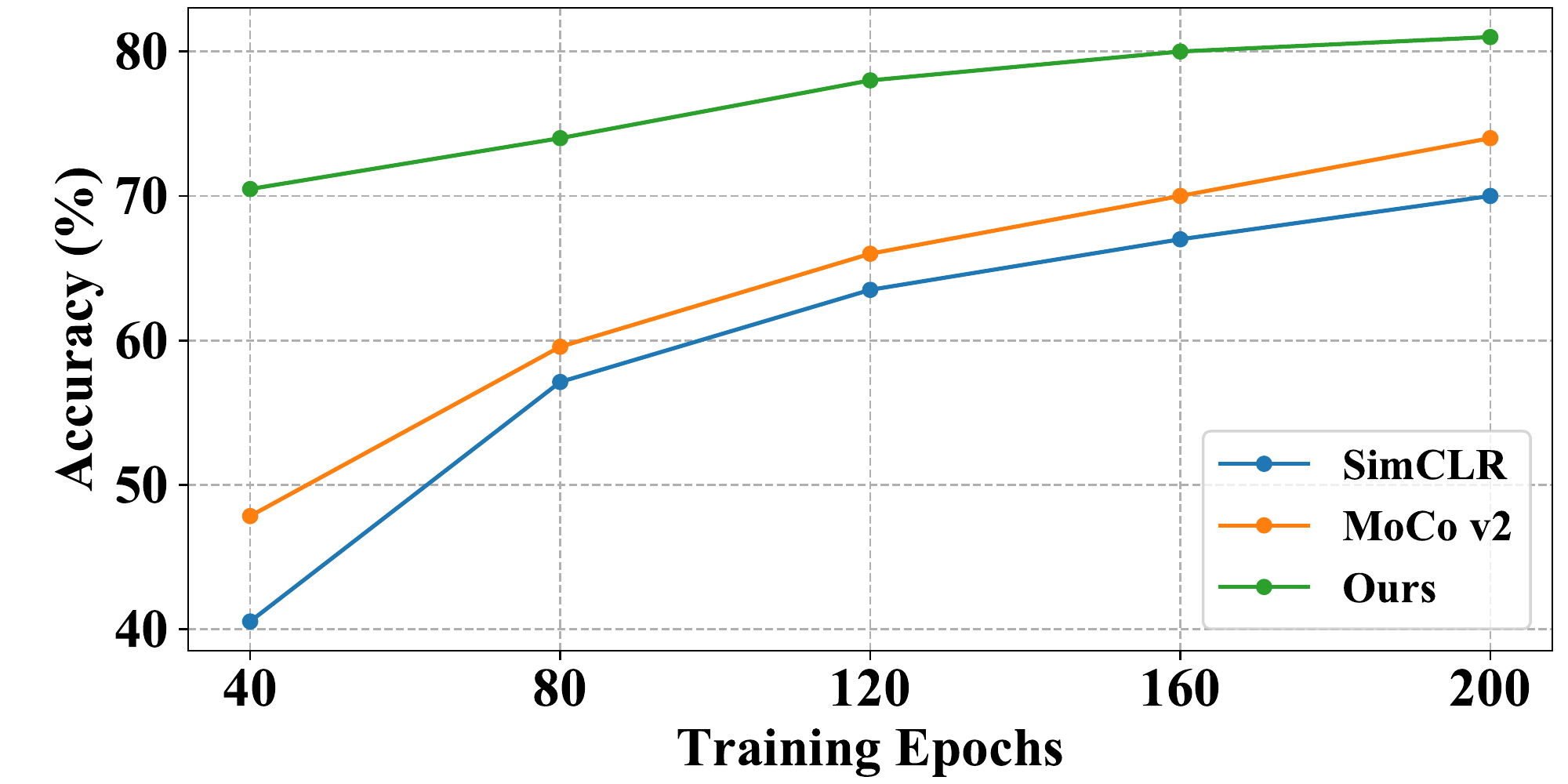}
	\caption{Results of ResNet-50 model on the ImageNet-100 dataset with linear evaluation protocol. The accuracy increases along with more of the total training epochs. Our method converges quickly. Note MoCo v2 costs around 160 epochs to reach the same level of output from our method in the 40th epoch.}
	\label{fig:trainingtime}
\end{figure}

\paragraph{Convergence}
We train unsupervised models on the ImageNet-100 dataset with different epochs and show the linear evaluation results in Figure \ref{fig:trainingtime}. Our method achieves decent results with a very small number of training epochs, while other contrastive learning methods require longer training time to reach the same accuracy. 

We explain that effective contrastive pairs are more frequent in our pretext tasks because of the split of input images. For exmaple, pairs in SimCLR are similar and are easy to recognize, leading to futile pairs. But the patches in the same cluster of our method come from different regions of the image, helpfully improve the quality of positive pairs.
 
\begin{table}
	\centering
	\begin{tabular}{l c c}
		\toprule
		Dataset & ImageNet-100 & ImageNet-10\%  \\
		\midrule
		SimCLR & 70.5 & 35.8 \\
		MoCo v2 & 74.7 & 38.3	 \\
		JigClu (Ours) & \textbf{80.9} & \textbf{44.3} \\
		\bottomrule
	\end{tabular}
	\vspace{0.1in}
	\caption{Linear evaluation results of ResNet-50 models on ImageNet-100 and ImageNet-10\% datasets. Our results are significantly better than those of other methods on small datasets.}
	\label{tab:linear2}
\end{table}

\subsection{Semi-supervised Learning}
We also finetune the unsupervised model under the semi-supervised setting on the ImageNet-1k dataset with 10\% and 1\% labels. The labels are still class-balanced, provided in \cite{simclr}. We finetune our model with a randomly initialized linear classifier on the labeled data. The results are summarized in Table \ref{tab:semi}.

Results of MoCo v2 are produced by us with the official model offered in \cite{mocogithub}. We train it for 200 epochs for fair comparison. Compared with state-of-the-art representation learning methods, we achieve better results with only half of the training batches. The results of semi-supervised learning further manifest the superiority of our method. Result of UDA is with higher accuracy because it is specially designed for semi-supervised learning and utilizes powerful RandAugment \cite{randaugment}. 

\begin{table}
	\centering
	\begin{tabular}{l c c c c}
		\toprule
		\multirow{3}{*}{Method} & \multicolumn{4}{c}{Label fraction} \\
		& \multicolumn{2}{c}{1\%} & \multicolumn{2}{c}{10\%} \\
		& Top-1 & Top-5 & Top-1 & Top-5 \\
		\midrule
		Supervised & 25.4 & 48.4 & 56.4 & 80.4 \\
		\midrule
		\multicolumn{5}{l}{\it{Methods using label-propagation:}} \\
		Pseudo-label \cite{pseudo} & - & 51.6 & - &82.4 \\
		Entropy-Min \cite{entropymin} & - & 47.0 & - & 83.4 \\
		S$^4$L-Rotation \cite{s4l} & - & 53.4 & - & 83.8 \\
		UDA* \cite{uda} & - & 68.8 & - & 88.5 \\
		\midrule
		\multicolumn{5}{l}{\it{Methods using unsupervised learning:}} \\
		NPID \cite{npid} & - & 39.2 & - & 77.4 \\
		PIRL \cite{pirl} & - & 57.2 & - & 83.8 \\
		MoCo v2 \cite{mocov2} & 34.5 & 62.2 & 61.1 & 83.9 \\
		JigClu (Ours) & \textbf{40.7} & \textbf{68.9} & \textbf{63.0} & \textbf{85.2} \\
		\bottomrule
	\end{tabular}
	\vspace{0.1in}
	\caption{Results of our pretrained model on the semi-supervised ImageNet classification tasks. Our method outperforms previous  unsupervised learning ones. * indicates using RandAugment \cite{randaugment}.}
	\label{tab:semi}
\end{table}

\subsection{Transfer Learning}

We apply our pretrained ResNet-50 models to other vision tasks to prove generalization of our ResNet-50 models trained on ImageNet.

\begin{table}
	\centering
	\begin{tabular}{l | c @{\hspace{0.1in}} c@{\hspace{0.1in}} c@{\hspace{0.1in}} c@{\hspace{0.1in}} c@{\hspace{0.1in}} c }
		\toprule
		Models & AP & AP50 & AP75 & APs & APm & APl \\
		\midrule
		MoCo v2 & 38.9 & 58.8 & 42.5 & 23.3 & 41.8 & \textbf{50.0} \\
		JigClu (Ours) & \textbf{39.3} & \textbf{59.4} & \textbf{42.5} & \textbf{23.6} & \textbf{42.5} & 49.7 \\
		\bottomrule
	\end{tabular}
	\vspace{0.1in}
	\caption{Results of Faster-RCNN R50-FPN models trained on COCO detection dataset with pretrained weights provided by unsupervised training on ImageNet.}
	\label{tab:coco}
\end{table}

\label{sec:coco}
\paragraph{Objection Detection} 
Following \cite{mocogithub}, we finetune our pretrained weights on the COCO detection dataset \cite{coco} with the Faster-RCNN R-50-FPN framework \cite{faster,fpn}. The results are summarized in Table \ref{tab:coco}. Our results are better than those of MoCo v2 pretrained weights. 
Our models learn comprehensive information including instance-level discrimination and location recognition, useful for the detection tasks.

\begin{table}
	\centering
	\begin{tabular}{l l@{\hspace{0.1in}}| @{\hspace{0.1in}} c@{\hspace{0.3in}} c}
		\toprule
		& Models & CIFAR-10 & CIFAR100 \\
		\midrule
		\multirow{3}{*}{\it{finetune}} & Rand init. & 88.4 & 61.6 \\
		& Supervised & 88.6 & 60.6 \\
		& JigClu (Ours) & \textbf{89.5} & \textbf{64.7} \\
		\midrule
		\multirow{2}{*}{\it{linear}} & Supervised & 62.5 & 41.0 \\
		& JigClu (Ours) & \textbf{68.8} & \textbf{45.0} \\
		\bottomrule
	\end{tabular}
	\vspace{0.1in}
	\caption{Results of ResNet-50 models trained on CIFAR-10 and CIFAR-100 datasets with different initialization.}
	\label{tab:cifar}
\end{table}

\paragraph{Image Classification}
We also apply our pretrained weights to CIFAR-10 and CIFAR-100 datasets. The classifiers of models are randomly initialized and backbones are initialized in different ways including random values, supervised training models on ImageNet, and unsupervised pretrained models on ImageNet in our JigClu task. 

The results are listed in Table \ref{tab:cifar}. We use finetuning and linear evaluation to test representation learned by JigClu. In the finetuning setting, the backbone and classifier are trained on the target dataset together. Our weights provide the best initialization for both CIFAR-10 and CIFAR-100 datasets. In the linear evaluation process, only the linear classifier is trained on new datasets. Our model is better than the supervised pretrained model on ImageNet, which demonstrates the generality of our learned representation.

\section{Analysis}
\label{sec:abl}

\subsection{Montage Input} 
\begin{figure}
	\centering
	\includegraphics[width=0.8\linewidth]{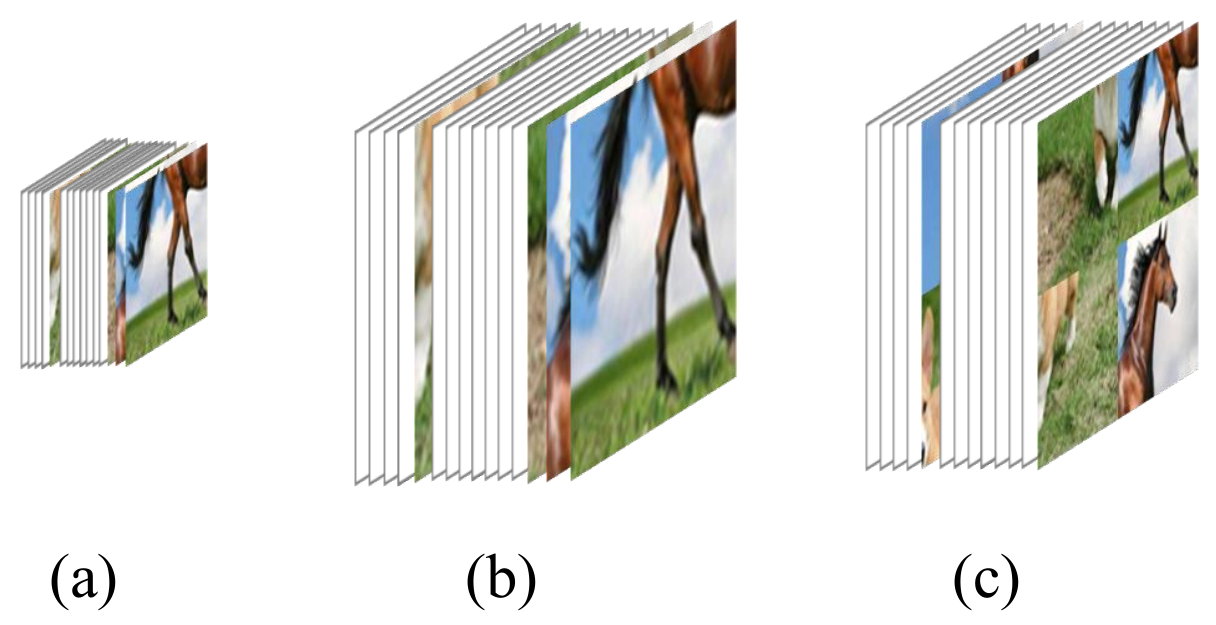}
	\caption{(a) Small size patches. (b) Scaled-up patches. (c) Montage images.}
	\label{fig:montage}
\end{figure}

We split every image in the batch into $m\times m$ patches and randomly permute them to form a new batch. The new batch used in our method consists of montage images as shown in Figure \ref{fig:montage}(c). 

Using montage images as input is better than directly using patches. 
On the one hand, if we do not scale up the patches (Figure \ref{fig:montage}(a)), the network is trained with small images, which greatly reduces the representation ability when dealing with high-resolution images. On the other hand, after we resize the patches into a larger shape (Figure \ref{fig:montage}(b)), the size of input batches is $m\times m$ times larger than the orginal one, which greatly increases demand of training resource. Putting aside the size of patches, straightly using patches as input may cause missing a lot of instance-level intra-image information. 

\begin{table}
	\centering
	\begin{tabular} {l | c c c}
		\toprule
		Input Format & Accuarcy (\%) & Time & Memory \\
		\midrule
		(a) Small-size Patch & 67.0 & 5.5h & 2700MB \\
		(b) Scaled-up Patch & 71.3 & 16.8h & 7300MB \\
		(c) Montage Image & 70.9 & 5.7h & 3000MB \\
		\bottomrule
	\end{tabular}
	\vspace{0.1in}
	\caption{Linear evaluation results of ResNet-18 models on the ImageNet-100 dataset. The scaled-up images lead to the best result, and yet cost much resource. Our proposed montage images achieve comparable results with scaled-up images, and only cost around 1/3 of the original training resource.}
	\label{tab:montage}
\end{table}

To prove the superiority of our proposed montage images, we conduct experiments on the ImageNet-100 dataset with ResNet-18 models. We choose ResNet-18 because it is much faster. Using scaled-up patches as input for ResNet-50 also causes the out-of-memory issue on our limited computing resource. We first train the models on the ImageNet-100 dataset, and evaluate them with a linear evaluation protocol. 

The results are reported in Table \ref{tab:montage}. 
Concluded from the table, using small patches as input is very fast, and yet leads to low performance. Scaled-up patches much improve result quality. But they cost too much resource during training. Using our montage images as input overcomes these limitations, attaining high performance on limited computing resource. Although the way to use montage images in our method is still primary, amazing results are yielded. This opens the door for future research of single-batch unsupervised methods with montage images.
 
\begin{figure}
	\centering
	\includegraphics[width=\linewidth]{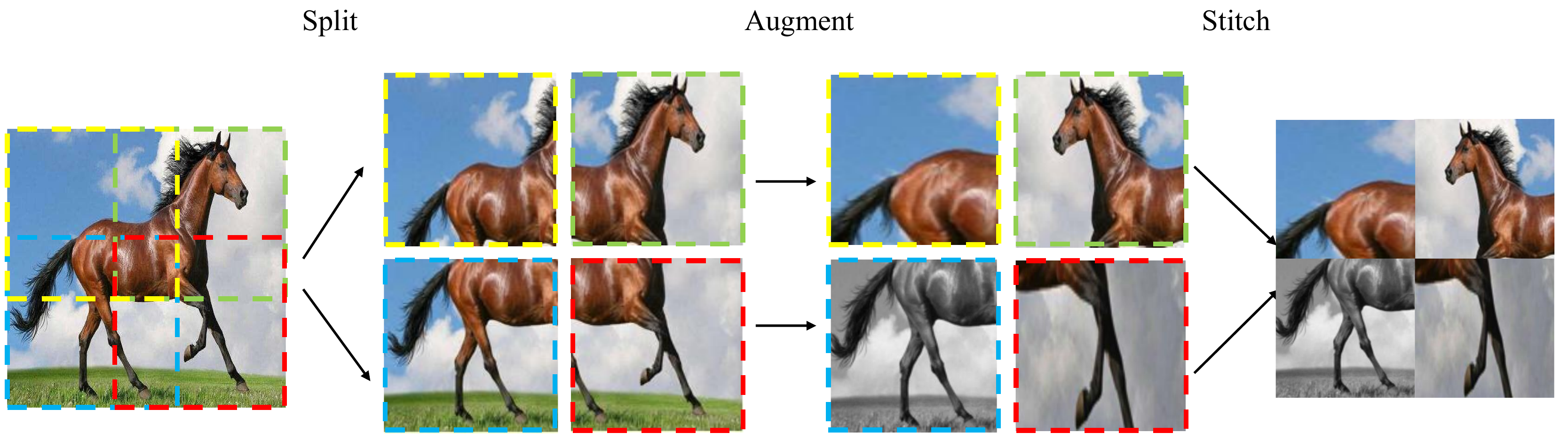}
	\caption{The position of augmentation used in our method. We augment every patch independently right after the split operation. In real cases, patches are mixed across images.}
	\label{fig:augment}
\end{figure}

\begin{table}
	\centering
	\begin{tabular}{l l}
		\toprule
		Augmentation position & Accuracy (\%) \\
		\midrule
		Aug before split & 3.7 \\	
		Split during aug & 39.3 \\
		Aug after split & \textbf{80.9} \\
		Aug after montage & Na \\
		\bottomrule
	\end{tabular}
	\vspace{0.1in}
	\caption{The linear evaluation results of ResNet-50 models on the ImageNet-100 dataset with different augmentation policies. Applying augmentation on patches individually produces the best results.}
	\label{tab:augment}
\end{table}

\subsection{Data Augmentation}
Data augmentation is very important in recent contrastive learning methods. We simply use the policy of MoCo v2 as our baseline augmentation. There is a split operation, which divides the image into $m\times m$ patches. We apply the baseline augmentation to every patch independently right after the split operation and before we perform the montage operation, as shown in Figure \ref{fig:augment}.
There are many other choices; but empirically we find this simple strategy suffices. We analyze the position of augmentation in this section. 

Using the augmentation after the montage operation is not feasible, because random-crop may cut off some patches from the montage image. 
The augmentation could be used on original images before the split operation. However, this would cause many problems. First, the clustering branch may learn the augmentation bias instead of image features, because patches from the same images use the same augmentation. Second, the location of patches is hard to learn in an image-agnostic manner, because the augmented images may be at any positions of the original images.
 
An improvement option is to use the split operation between transform. For example, we could first crop the original images, and then split the cropped image into patches. These patches are further transformed by other operations such as color jitterring. This strategy only partially solves previous problems. Our augmentation right on patches tackles these issues.

We experiment with different augmentation ways, and report the results in Table \ref{tab:augment}. The ResNet-50 models are unsupervisedly pretrained on ImageNet-100 dataset with different augmentation policies, and the linear evaluation results are used to measure learning of representation. The model learns nothing when images are augmented before the split operation. Bringing forward the crop operation helps the model learn some useful information to produce non-trivial results. Applying augmentation to every patch is clearly a decent choice. It obtains high-quality results.

\subsection{Split Operation}

\begin{table}
	\centering
	\begin{tabular}{c @{\hspace{0.2in}}| @{\hspace{0.2in}}c@{\hspace{0.2in}} c @{\hspace{0.2in}} c}
		\toprule
		$m$ & 2 & 3 & 4\\
		\midrule
		Accuracy (\%) & 80.9 & 74.7 & 70.1\\
		\bottomrule
	\end{tabular}
	\vspace{0.1in}
	\caption{The linear evaluation results of ResNet-50 models on the ImageNet-100 dataset with different $m$. When $m$ is larger than 2, the performance decreases because of the increased difficulty.}
	\label{tab:patches}
\end{table}

\paragraph{Number of Patches}
We split every image into $m\times m$ patches, where the choice of $m$ is highly related to our task. The minimum value for $m$ is 2. So we start from 2 to find an optimal $m$. The results are listed in Table \ref{tab:patches}. 

We report the linear evaluation results of unsupervised learning ResNet-50 models on the ImageNet-100 datasets. It is obvious to conclude that the result of $m=3$ is worse than $m=2$. And we do not try larger $m$s. This result shows that setting $m=3$ already makes the model difficult to learn because of the small input size ($224\times 224$). In this case, it becomes a problem to distinguish among these many patches inside one montage image. Without further notes, we use $m=2$ in all our experiments.

\begin{figure}
	\centering
	\includegraphics[width=0.8\linewidth]{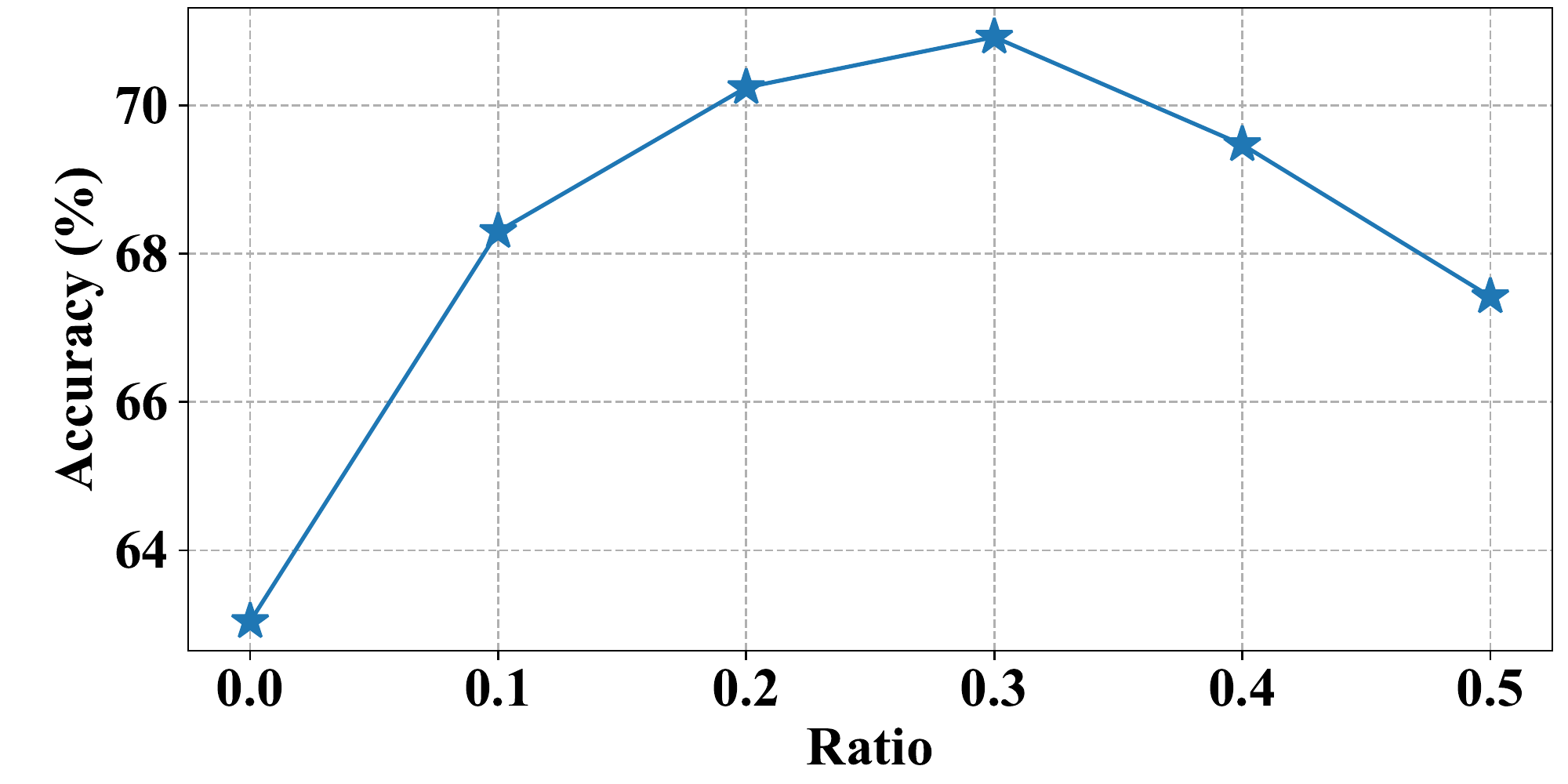}
	\caption{Accuracy in terms of different overlap ratios of adjacent patches. The $x$-axis is the length of overlapping regions, which is measured in terms of the percentage of image size. When $x=0$, the patches are completely separated without overlap.}
	\label{fig:across}
\end{figure}

\paragraph{Overlapped Region Size}
The size of overlap between adjacent patches also influences our method. We conduct experiments to find an optimal size for the overlapped regions. We train ResNet-18 on the ImageNet-100 dataset with our proposed unsupervised pretext task and use the linear evaluation to measure learning results. We use $m=2$ in our experiments. 

The results are summarized in Figure \ref{fig:across}. Concluded from the figure, when there is no overlap between patches, the model does not learn effective features, because it is hard to distinguish among patches from the same image without any ideas how they are overlapped. When the overlap between patches becomes large, the result quality also drops. In this condition, patches from the same image may be very similar to each other and are easy to recognize, leading to reduction of effective positive pairs. 

We find that using 0.3 of the original image's side length produces the best results. This ratio achieves a good balance of difficulty and efficiency for positive pairs. All our experiments are trained using this ratio.

\begin{table}
	\begin{tabular}{c c r}
		\toprule
		Clustering branch & Location branch & Accuracy (\%) \\
		\midrule
		\ding{52} & & 65.1 \\
		& \ding{52} & 3.2 \\
		\ding{52} & \ding{52} & 66.4 \\
		\bottomrule
	\end{tabular}
	\vspace{0.1in}
	\caption{Ablation study of the two branches. The results is measures by the linear evaluation protocol of ResNet-50 models on the ImageNet-1k dataset.}
	\label{tab:branch}
\end{table}

\subsection{Importance of the Two Branches}
The proposed pretext task is solved by the two branches: clustering branch and location branch. Each branch has a loss function. The clustering branch is supervised by a contrastive-like loss, aiming to cluster patches from the same original images. This branch dominates the training of models. Both instance- and image-level information is learned from this branch. The location branch is supervised by a classification loss, which predicts the position of every patch in an image-agnostic manner. This branch assists the clustering branch with more detailed location information. 

We train branches separately and summarize the results in Table \ref{tab:branch}. The results are measured on linear evaluation of unsupervised training with ResNet-50 models on the ImageNet-1k dataset.We can observe from the table that only training with the location branch leads to trivial accuracy. It reflects that the location information cannot be effectively learned from such complex montage input individually. However, the location branch is a nice auxiliary for the clustering branch. Joint training of both branches achieves the best results.

\section{Conclusion}
In this paper, we have proposed a novel Jigsaw Clustering pretext task/method, taking the advantage of both contrastive learning and previous handcrafted pretext tasks. Models trained with our method can learn both intra- and inter-images information with a single batch during training. Our method outperforms previous single-batch ones by large margins, and achieves comparable results with dual-batch methods with only half of the training batches. Our method naturally applies to other tasks. 

Our work manifests, intriguingly, that single-batch methods have the potential to be in par with or even outperform dual-batch ones. We believe this line is worth further study. New applications can be expected.

{\small
\bibliographystyle{ieee_fullname}
\bibliography{egbib}
}

\end{document}